\PassOptionsToPackage{main=english,russian}{babel}
\PassOptionsToPackage{T2A}{fontenc}
\documentclass[onecolumn]{vision_guided_chunk}

\usepackage{lipsum}
\usepackage{pdfpages}
\usepackage{colortbl}
\usepackage{tabulary}
\usepackage{longtable}
\usepackage{array}
\usepackage{placeins}
\usepackage{tablefootnote}
\usepackage{tcolorbox}
\usepackage{wrapfig}
\usepackage{listings}

\usepackage{breqn} 

\usepackage{pifont}
\definecolor{deeppurple}{HTML}{9e02f7}
\definecolor{forestgreen}{HTML}{2e7d43}
\usepackage{multirow}
\usepackage{tabularx}

\usepackage[utf8]{inputenc}

\usepackage{arydshln}

\usepackage{xspace} 
\usepackage{makecell} 
\usepackage{multirow}

\usepackage[framemethod=TikZ]{mdframed}
\usepackage{tcolorbox}
\usepackage{multicol}
\usepackage{dblfloatfix}

\newtcolorbox{mybox}[2][]{
  colback=white, 
  colframe=lightblue,
  fonttitle=\bfseries,
  coltitle=black,  
  title=#2, 
  #1
}

\definecolor{ayad}{RGB}{148, 156, 229} 
\definecolor{ayadsymbol}{RGB}{76, 110, 230} 
\definecolor{lightblue}{RGB}{211, 227, 252} 
\definecolor{bgblue}{RGB}{247, 250, 255} 
\newcommand*\colourcheck[1]{%
  \expandafter\newcommand\csname #1check\endcsname{\textcolor{#1}{\ding{52}}}%
}
\newcommand*\colourcross[1]{%
  \expandafter\newcommand\csname #1cross\endcsname{\textcolor{#1}{\ding{55}}}%
}
\DeclareSymbolFont{extraup}{U}{zavm}{m}{n}
\DeclareMathSymbol{\vardiamond}{\mathalpha}{extraup}{87}
\definecolor{ayadsymbol}{RGB}{76, 110, 230} 

\colourcheck{blue}
\colourcheck{green}
\colourcheck{red}

\colourcross{blue}
\colourcross{green}
\colourcross{red}
\usepackage{nicematrix}
\usepackage{comment}
\usepackage{amssymb}

\setcitestyle{number,square}

\title{Vision-Guided Chunking Is All You Need: Enhancing RAG with Multimodal Document Understanding}

\multiauthors

\author{
    name={Vishesh Tripathi$^\circ$}
}

\author{
    name={Tanmay Odapally$^\circ$}
}

\author{
    name={Indraneel Das$^\circ$}
}

\author{
    name={Uday Allu$^\dagger$}
}

\author{
   name={Biddwan Ahmed$^\dagger$}
}

\affiliations{
    AI Research Team, Yellow.ai}

\date{\today}
\abstract{
Retrieval-Augmented Generation (RAG) systems have revolutionized information retrieval and question answering, but traditional text-based chunking methods struggle with complex document structures, multi-page tables, embedded figures, and contextual dependencies across page boundaries. We present a novel multimodal document chunking approach that leverages Large Multimodal Models (LMMs) to process PDF documents in batches while maintaining semantic coherence and structural integrity. Our method processes documents in configurable page batches with cross-batch context preservation, enabling accurate handling of tables spanning multiple pages, embedded visual elements, and procedural content. We evaluate our approach on our internal benchmark dataset of diverse PDF documents, demonstrating improvements in chunk quality and downstream RAG performance. Our vision-guided approach achieves better quantitative performance on our internal benchmark compared to traditional vanilla RAG systems, with qualitative analysis showing better preservation of document structure and semantic coherence.
}

\renewcommand{\FONTauthors}{\bfseries\Large}

\begin{document}

\footnotetext[1]{$^\circ$First authors}
\footnotetext[2]{$^\dagger$Senior authors}
\footnotetext[3]{\{vishesh.tripathi1,tanmay.odapally1,indraneel,uday,biddwan\} @yellow.ai}

\section{Introduction}

Retrieval-Augmented Generation (RAG)~\citep{lewis2020retrieval} has emerged as an important paradigm for enhancing large language models with external knowledge sources. The effectiveness of RAG systems fundamentally depends on the quality of document chunking - the process of segmenting documents into coherent, retrievable units. Traditional approaches rely on simple text extraction followed by rule-based or sliding-window chunking~\citep{carbonell1998use}, which often fails to preserve semantic coherence and structural relationships in complex documents.

Modern documents, particularly technical manuals, research papers, and business reports, contain rich multimodal content including tables, figures, diagrams, and multi-page structures that span across page boundaries. These elements are crucial for understanding but are often lost or fragmented by conventional text-only processing methods.

Recent advances in Large Multimodal Models (LMMs)~\citep{yin2023survey} present an opportunity to revolutionize document processing by leveraging both visual and textual understanding. We propose a novel chunking methodology that processes PDF documents using LMMs in configurable batches while maintaining contextual continuity across batch boundaries.

\subsection{Contributions}

Our main contributions are:

\textbf{Multimodal Batch Processing Framework}: A novel approach to process PDF documents in page batches using LMMs, enabling better handling of complex layouts and visual elements.

\textbf{Context Preservation Mechanism}: A systematic method for maintaining semantic continuity across batch boundaries, solving the context loss problem in large document processing.

\textbf{Structural Integrity Preservation}: Techniques for maintaining table structures, step-by-step procedures, and multi-page content relationships.

\textbf{Comprehensive Evaluation and Benchmarking}: Development of a new benchmark for multimodal document chunking evaluation, with validation using Gemini-2.5-Pro as the chunking model and GPT-4.1 as the summarization model on diverse document types with downstream RAG performance assessment."

\section{Related Work}

\subsection{Traditional Document Chunking}
Traditional RAG systems employ various chunking strategies. Fixed-size chunking segments documents into fixed-length pieces, often losing semantic boundaries and breaking coherent concepts across multiple chunks. Sentence-based chunking uses sentence boundaries as natural breakpoints but ignores document structure and hierarchical relationships between content sections. Paragraph-based chunking preserves paragraph structure but struggles with complex layouts, tables, and multi-page content that spans across traditional paragraph boundaries. Semantic chunking attempts to identify semantic boundaries using natural language processing techniques, but relies solely on text-only features and fails to capture visual and structural elements that are crucial for document understanding.

\subsection{Multimodal Document Understanding}

Recent work in multimodal document understanding has made significant advances across several areas. Document layout analysis using vision transformers~\citep{dosovitskiy2020image} has enabled better understanding of document structure, including detection of headers, paragraphs, and reading order. Pre-trained models like LayoutLM~\citep{xu2020layoutlm} and LayoutLMv2~\citep{xu2021layoutlmv2} have improved the ability to process structured data within documents. Large-scale vision foundation models like InternVL~\citep{chen2024internvlscalingvisionfoundation} have advanced generic visual-linguistic understanding capabilities for complex document processing tasks. Table detection and extraction using multimodal models has improved the ability to process structured data within documents, with specialized approaches for contextualizing tabular data in RAG systems~\citep{allu2024extractioncontextualisingtabulardata}, though challenges remain for tables spanning multiple pages. Modern document conversion toolkits like Docling~\citep{livathinos2025doclingefficientopensourcetoolkit} have provided efficient open-source solutions for AI-driven document processing workflows. Figure captioning and visual question answering for documents~\citep{mathew2021docvqa} has enhanced the extraction of information from charts, diagrams, and images embedded within text. End-to-end document understanding with unified multimodal architectures has shown promise in creating comprehensive document representations that combine visual and textual information.

\subsection{RAG System Optimization}
Prior work on improving RAG systems has focused on several key areas. Better retrieval mechanisms, including dense retrieval~\citep{karpukhin2020dense} and hybrid search approaches~\citep{li2021hybrid}, have improved the accuracy of finding relevant information. Query expansion and reformulation techniques~\citep{nogueira2019passage} have enhanced the matching between user queries and document content. Re-ranking and filtering strategies have helped prioritize the most relevant retrieved chunks for generation. Multi-hop reasoning approaches~\citep{yang2018hotpotqa} have enabled more complex question answering that requires combining information from multiple sources. Techniques like vision-RAG~\citep{chen2024visrag} and Video-RAG~\citep{zhang2024videorag} have allowed the framework to use multimodality. However, limited attention has been paid to improving the fundamental chunking process using multimodal understanding, which represents a significant gap in the current literature.

\section{Methodology}

Traditional document chunking approaches face several fundamental limitations when processing complex PDF documents. Fixed-size and sliding-window methods fragment coherent content across chunk boundaries, breaking multi-page tables, step-by-step procedures, and cross-referential relationships. Text-only extraction completely ignores visual elements such as figures, charts, and document layout structure, which often contain critical information for understanding. Furthermore, conventional approaches fail to preserve semantic relationships that span across page boundaries, resulting in contextually incomplete chunks that hinder effective retrieval. The hierarchical organization of documents—including nested sections, subsections, and procedural sequences—is typically lost, making it difficult for RAG systems to understand the logical flow and dependencies within the document. These limitations become particularly pronounced in technical documents, financial reports, and regulatory filings where structural integrity and visual elements are essential for accurate interpretation.

\subsection{Problem Formulation}
Let $D$ be a PDF document with $n$ pages: $D = \{p_1, p_2, \ldots, p_n\}$. Traditional text-only chunking produces chunks $C = \{c_1, c_2, \ldots, c_m\}$ where each chunk $c_i$ contains only textual content extracted from pages.

Our multimodal approach processes $D$ in batches $B = \{B_1, B_2, \ldots, B_k\}$ where each batch $B_i$ contains up to $b$ consecutive pages (typically $b = 4$):

$$B_i = \{p_j : (i-1) \cdot b + 1 \leq j \leq \min(i \cdot b, n)\}$$

This ensures that batch $i$ contains pages from $(i-1) \cdot b + 1$ to $\min(i \cdot b, n)$, properly handling cases where the document length is not evenly divisible by the batch size.

For each batch $B_i$, we generate contextually-aware chunks $C_i$ using a Large Multimodal Model $M$:

$$C_i = M(B_i, \text{context}_{i-1}, \text{prompt})$$

where $\text{context}_{i-1}$ represents relevant context from previous batches.

\subsection{Multimodal Batch Processing}

\begin{figure*}[!t]
     \centering
     \includegraphics[width=1\textwidth]{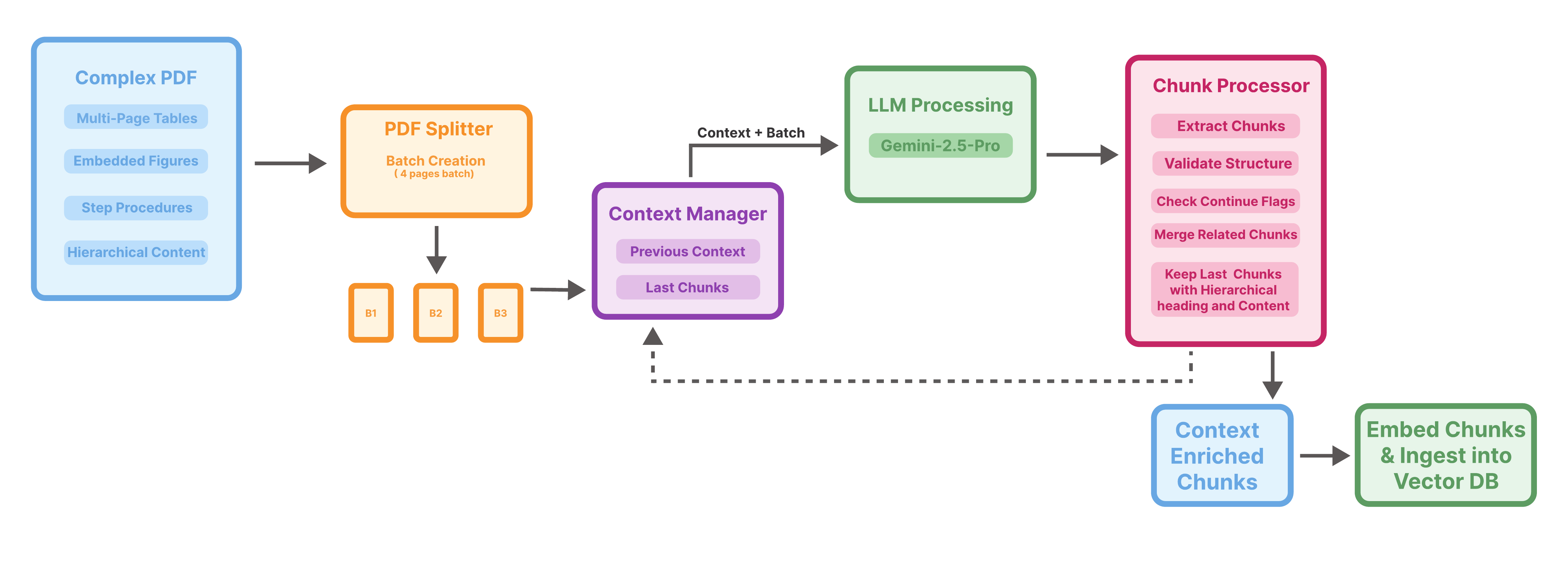}
     \caption{Multimodal Document Chunking Architecture: Our framework processes PDF documents in configurable page batches using Large Multimodal Models (LMMs), maintaining cross-batch context through continuation flags and hierarchical heading structures. The system preserves semantic coherence across page boundaries while handling complex elements like multi-page tables, embedded figures, and procedural content.}
     \label{fig:multimodal_chunking_pipeline}
\end{figure*}

Our multimodal batch processing framework, depicted in Figure~\ref{fig:multimodal_chunking_pipeline}, addresses the fundamental limitations of traditional text-only chunking by leveraging the visual understanding capabilities of Large Multimodal Models. Documents are split into batches of $b$ pages, with each batch processed through our vision-guided pipeline that maintains contextual relationships across page boundaries.

\subsubsection{Batch Creation}
Documents are split into batches of $b$ pages. The batching process ensures that related content spanning multiple pages can be processed together, maintaining contextual relationships that would be lost in traditional page-by-page processing.

For a document with $n$ pages, the number of batches $k$ is calculated as:
$$k = \lceil \frac{n}{b} \rceil$$

Each batch $B_i$ contains at most $b$ pages, with the final batch potentially containing fewer pages if $n$ is not divisible by $b$.

\subsubsection{Context Preservation}
To maintain continuity across batches, we implement a context mechanism that includes previous context, the final chunks from the previous batch to handle content spanning batch boundaries, and maintained heading hierarchy to ensure consistent organization.

The context for batch $B_i$ is constructed as:
$$\text{context}_i = \{\text{last\_chunk}_{i-1}, \text{heading\_hierarchy}_{i-1}\}$$

This context mechanism ensures that information from previous batches informs the processing of subsequent batches, preventing the loss of semantic relationships across batch boundaries.

\subsection{Intelligent Chunk Generation}

\subsubsection{Hierarchical Heading Structure}

We enforce a consistent 3-level heading hierarchy throughout the document processing based on empirical analysis of our document corpus. Our evaluation showed that 2-level hierarchies lost important contextual granularity for complex documents, while 4+ levels introduced unnecessary fragmentation that degraded retrieval performance. The 3-level structure strikes an optimal balance between semantic granularity and retrieval efficiency. Level 1 headings represent the document or product title with full details including location and context information. Level 2 headings capture major sections such as "Features", "Procedures", or "Specifications". Level 3 headings identify specific subtopics including "Step 1", "Table Row", or detailed subsections.

This hierarchical structure ensures that each chunk maintains its contextual position within the overall document structure, enabling better retrieval and understanding during the RAG process.

The whole prompt used can be found in the Appendix.

\subsubsection{Content Preservation Rules}
Critical rules for maintaining document integrity include several key principles. Step preservation ensures that all numbered steps or procedures remain in the same chunk, preventing fragmentation of instructional content. Table integrity maintains that each table row becomes a separate chunk while preserving headers for context. List continuity keeps related list items together as coherent units. Multi-page structures are properly merged when content spans across page boundaries.

These rules are implemented through careful parsing of the multimodal model output and post-processing validation to ensure compliance with structural requirements.

\subsubsection{Continuation Flags}
Each chunk is tagged with a continuation flag to enable intelligent post-processing. The flag system uses three categories: \texttt{[CONTINUES]True[/CONTINUES]} for chunks that continue from previous content, \texttt{[CONTINUES]False[/CONTINUES]} for chunks representing new content, and \\ \texttt{[CONTINUES]Partial[/CONTINUES]} for uncertain continuation relationships.

This tagging system enables automated merging of related content during post-processing, ensuring that semantically related chunks are appropriately combined while maintaining proper boundaries between distinct topics.

\subsection{Mathematical Framework for Retrieval}
In the retrieval phase, given a query $q$, we compute similarity scores using cosine similarity, as a method to compare similarity between sentences ~\citep{reimers2019sentence}:
$$\text{sim}(q, c_i) = \frac{E(q) \cdot E(c_i)}{||E(q)|| \cdot ||E(c_i)||}$$

where $E(\cdot)$ represents the embedding function that maps text to dense vector representations.

The top-K chunks are selected as:
$$K = \{c_{i_1}, c_{i_2}, \ldots, c_{i_k}\} \text{ where } \text{sim}(q, c_{i_j}) \geq \text{sim}(q, c_{i_{j+1}})$$

The enhanced chunks from our multimodal processing provide richer context for similarity computation, leading to improved retrieval performance compared to traditional text-only chunks.

\section{Implementation Details}

\subsection{System Architecture}
Our implementation consists of several key components working in coordination. The PDF Processor handles document downloading and batch creation, managing the splitting of large documents into processable units. The Multimodal Interface manages communication with LMMs including Gemini-2.5-Pro, handling API calls and response processing. The Context Manager maintains cross-batch context and heading hierarchies, ensuring continuity across processing boundaries. The Chunk Processor extracts and validates chunks from model responses, applying the continuation rules and structural requirements. Finally, the Database Integration component prepares chunks for vector storage and retrieval in the RAG system.

\subsection{Model Configuration}

We experiment with the current state-of-the-art multimodal model for our evaluation. Gemini-2.5-Pro represents Google's latest multimodal model with enhanced document understanding capabilities, particularly strong in handling complex layouts and visual elements.

The model is configured with low temperature settings ($T = 0.1$) to ensure consistent and reliable chunk generation, minimizing variability in output structure while maintaining the quality of content extraction.

\subsection{Prompt Engineering}
Our prompt design incorporates several critical elements for effective chunk generation. Detailed chunking instructions with priority rules guide the model in making decisions about content segmentation. Examples of proper heading hierarchy provide concrete templates for maintaining consistent structure across batches. Special handling instructions for tables, steps, and multi-page content ensure that complex structural elements are processed correctly. Context integration guidelines specify how previous batch information should influence current processing decisions.

The prompt engineering process involved iterative refinement based on initial results, with particular attention to edge cases involving table structures and procedural content that spans multiple pages.

\section{Experiment}

\subsection{Setup}

We evaluate our vision-guided chunking approach within a complete RAG pipeline to demonstrate the impact of improved document parsing on downstream performance. Our experimental setup consists of two main components:

\textbf{Vision-Guided Chunking Pipeline}: We employ our proposed multimodal batch processing framework using Gemini-2.5-Pro to process PDF documents in batches of 4 pages with context preservation mechanisms. The chunking pipeline generates semantically coherent chunks while maintaining document structure, table integrity, and cross-page relationships as described in Section 3.

\textbf{RAG System Configuration}: Following chunk generation, we construct a standard RAG pipeline where document chunks are embedded using OpenAI text-embedding-3-small~\citep{neelakantan2022text} and stored in an Elasticsearch~\citep{elasticsearch} vector database. For retrieval, we use top-k similarity search (k=10) to identify relevant chunks for each query. Post-retrieval, we employ GPT-4.1 for response generation. We use GPT-4.1-mini for evaluation, as this task does not require the complexity of larger models and allows for efficient evaluation while maintaining response quality.

\begin{table}[h]
\centering
\begin{tabular}{lc}
\toprule
\textbf{Chunking Method} & \textbf{Accuracy} \\
\midrule
Vanilla RAG (Fixed-size chunking) & 0.78 \\
Vision-Guided RAG (Our approach) &  0.89\\
\bottomrule
\end{tabular}
\caption{RAG System Performance Comparison}
\label{tab:rag_performance}
\end{table}

\subsection{Dataset}

We curated a comprehensive dataset comprising documents from multiple domains to evaluate the effectiveness of our vision-guided chunking approach. The dataset includes technical manuals, financial reports, research publications, regulatory documents, and business presentations, ensuring diverse document structures and complexity levels.

Our dataset is strategically designed to test various challenging aspects of document understanding:

\textbf{Document Structure Complexity}: Documents containing multi-level hierarchical organization with our enforced 3-level heading structure (Document Title > Section Heading > Subsection Heading), nested tables spanning multiple pages, embedded figures and diagrams, and cross-references and footnotes.

\textbf{Content Diversity}: Technical procedural instructions with step-by-step workflows, financial data with complex tabular structures, regulatory compliance documentation, research papers with mathematical formulations, and business reports with mixed content types.

\textbf{Visual Elements}: Multi-page tables requiring header preservation, flowcharts and process diagrams, embedded charts and graphs, and complex layouts with multi-column text.

For evaluation, we manually developed a comprehensive set of realistic queries that test both simple factual retrieval and complex analytical reasoning. These queries are designed to assess:
\begin{itemize}
\item \textbf{Factual Information Extraction}: Direct retrieval of specific data points, figures, and statements
\item \textbf{Cross-Table Analysis}: Queries requiring information synthesis across multiple table sections
\item \textbf{Procedural Understanding}: Questions about step-by-step processes and instructions
\item \textbf{Multi-Section Reasoning}: Complex queries requiring integration of information from different document sections
\item \textbf{Structural Comprehension}: Questions that test understanding of document hierarchy and organization
\end{itemize}
The query distribution make sures we have a balanced coverage across different difficulty levels and content types, providing a robust benchmark for evaluating RAG system performance improvements through enhanced chunking quality.

\subsection{Evaluation Metrics}

We employ a comprehensive evaluation framework that assesses both component-level and end-to-end system performance:

\textbf{RAG Performance Metrics}: We evaluate the complete RAG pipeline using accuracy as the primary metric, where GPT-4.1-mini serves as an automated judge~\citep{zheng2023judging} to validate answer correctness. This lightweight evaluation approach is suitable for this task while maintaining reliability and efficiency in assessment. We have added the prompt used for RAG validation in Appendix A.2.

\textbf{Chunk Quality Analysis}: We conduct manual qualitative analysis of generated chunks to assess semantic coherence, structural preservation, and information completeness. This includes evaluating the retention of table structures, cross-page relationships, and hierarchical document organization compared to traditional chunking methods.

The evaluation framework ensures comprehensive assessment of both the technical improvements in chunking quality and the practical impact on downstream RAG performance across diverse document types and query complexities.

\section{Results and Discussion}

\subsection{Chunk Quality Analysis}

Manual inspection of chunks generated by our vision-guided approach reveals significant improvements in semantic coherence and structural preservation compared to traditional text-only methods. Our approach successfully maintains table integrity across page boundaries, preserves procedural instruction sequences, and retains hierarchical document organization that is often lost in conventional chunking approaches.

Key qualitative improvements observed include: (1) Complete preservation of multi-page  tables with proper header repetition, (2) Intact cross-reference systems linking footnotes to relevant table cells, (3) Maintained procedural sequences in regulatory compliance sections, and (4) Proper handling of nested organizational structures in complex  documents. Selected examples of superior chunk quality are provided in Appendix for detailed comparison.

\subsection{RAG System Performance}

Evaluation of the complete RAG pipeline demonstrates substantial improvements when using our vision-guided chunks compared to traditional approaches. Table~\ref{tab:rag_performance} presents the accuracy results across our curated document dataset.

The improvement in accuracy demonstrates the benefits of better document parsing on downstream RAG performance. Beyond quantitative improvements, our vision-guided chunking method significantly enhances chunk observability - the ability to understand, trace, and validate the content within each chunk - and overall system explainability. This improved observability stems from our hierarchical heading structure and context preservation mechanisms, which provide clear semantic boundaries and maintain document relationships that are often lost in traditional chunking approaches.

Notably, our analysis reveals a substantial difference in chunking granularity between approaches. Traditional vanilla parsing generated significantly fewer chunks due to its rigid text-extraction limitations and fixed-size constraints. In contrast, our vision-guided approach produced approximately 5 times more chunks, demonstrating the language model's intelligence in creating more systematic and contextually appropriate segmentation. This increased granularity enables more precise retrieval by allowing the system to identify and extract specific, relevant information rather than retrieving large, heterogeneous text blocks that may contain both relevant and irrelevant content.

The improved performance is attributed to our approach's ability to maintain semantic coherence across page boundaries, preserve critical structural information, and generate contextually rich chunks that enable more accurate retrieval and response generation.GPT-4.1-mini's evaluation confirms that responses generated using our vision-guided chunks are more accurate, complete, and structurally coherent compared to those produced by vanilla RAG systems.

\section{Future Work}

Several promising directions emerge from this research that warrant further investigation and development.

\textbf{Advanced Multimodal Integration}: Future work could explore deeper integration of visual elements through improved figure understanding, better mathematical formula processing, and better handling of complex diagrams and charts. Additionally, investigating the potential of newer multimodal architectures and their specific strengths in document understanding could yield further improvements.

\textbf{Scalability and Optimization}: Research into more efficient batch processing strategies, reduced computational costs through model optimization, and real-time processing capabilities would improve the practical applicability of the approach. This includes investigating techniques for adaptive batch sizing based on document complexity and content density.

\textbf{Standardized PDF Benchmarking Dataset}: Through our extensive evaluation process, we identified a significant gap in the availability of reliable, comprehensive PDF benchmarking datasets for document understanding tasks.

\section{Challenges and Limitations}

While our multimodal chunking approach demonstrates significant improvements over traditional methods, several challenges remain that require further investigation. The most prominent limitation occurs when processing extremely complex tables that span 8-9 pages or more, where maintaining consistent column alignment and semantic relationships across such extensive structures becomes increasingly difficult for current LMMs to handle reliably. Additionally, highly complex figures such as intricate flowcharts, multi-layered technical diagrams, and dense statistical charts with embedded sub-elements present ongoing challenges for accurate extraction and description, as these visual elements often contain nuanced information that requires domain-specific understanding beyond current multimodal capabilities. Furthermore, the computational cost and processing time increase substantially with document complexity and batch size, potentially limiting real-time applications, while the approach's effectiveness remains dependent on the underlying LMM's vision capabilities, which may vary across different model architectures and continue to evolve rapidly.

\section{Conclusion}

We present a novel multimodal approach to document chunking that significantly improves upon traditional text-only methods for RAG systems. By leveraging Large Multimodal Models with batch processing and context preservation, our method successfully handles complex document structures, multi-page content, and visual elements while maintaining semantic coherence and structural integrity. The approach demonstrates the potential of multimodal AI in enhancing fundamental components of RAG systems, moving beyond simple text extraction to comprehensive document understanding. The systematic evaluation across diverse document types validates the generalizability and robustness of the method. As multimodal models continue to improve and become more cost-effective, we expect this methodology to become increasingly practical for production RAG applications. Our work opens new avenues for document understanding in information retrieval systems and provides a foundation for future research in multimodal RAG architectures. We encourage researchers to build upon our open-source framework, explore domain-specific applications, and further advance the integration of visual understanding in document processing systems.

\bibliography{addon}

\clearpage
\newpage
\appendix

\section{Appendix}

\subsection{Complete Chunking Prompt}

\vspace{1em}

\begin{mdframed}[
    frametitle={Multimodal Document Chunking Prompt},
    frametitlebackgroundcolor=lightblue!50,
    backgroundcolor=lightblue!10,
    linecolor=lightblue!50,
    linewidth=1pt,
    roundcorner=3pt,
    innertopmargin=10pt,
    innerbottommargin=10pt,
    innerleftmargin=10pt,
    innerrightmargin=10pt
]

{\small
{\itshape Extract text from the provided PDF and segment it into contextual chunks for knowledge retrieval while following these comprehensive requirements:}

\vspace{0.3em}
\textbf{EXTRACTION PHASE}

Process the PDF page by page, make sure you go through each page, don't skip any page, extracting all content while:

\begin{enumerate}
\item Read all data content carefully and understand the structure of the document.

\item Infer logical headings and topics based on the content itself.

\item Always generate a 3-level heading structure for every chunk:
   \begin{itemize}
   \item First-level heading = Document or product title
   \item Second-level heading = the major section inside the document
   \item Third-level heading = the specific subtopic within that section 
   \item Important: if heading is missing, inherit from the parent heading level. Use your best judgment to logically assign headings based on the content and fully—never paraphrase or shorten. The headings hierarchy should always follow this pattern: Main Title > Section Title > Chunk Title for headings.
   \end{itemize}

\item SKIP TABLE OF CONTENTS AND INDEXES: Do not create chunks from tables of contents or indexes.

\item Do not include page headers, footers and page numbers in the chunks.

\item Do not create or extract chunks from LAST CHUNKS. Use it only as guidance for heading inference. All chunks must originate directly from the image.

\item DO NOT alter, paraphrase, shorten, or skip any content. All text, formatting, and elements must remain exactly as in the original Image and present in the output.
\end{enumerate}

\vspace{0.3em}
\textbf{CRITICAL: STEP/LIST CHUNKING RULES - HIGHEST PRIORITY}

KEEP ALL RELATED CONTENT TOGETHER - This is the highest priority rule:
\begin{itemize}
\item \textbf{NEVER EVER split numbered steps, instructions, or procedures across different chunks}
\item \textbf{ALL steps in a set of instructions MUST stay together in the same chunk}
\item \textbf{ALL items in a numbered or bulleted list MUST stay together in one chunk}
\item If a list or set of steps spans multiple images, they MUST still be kept in a single chunk
\item If a list or steps continue from a previous batch, merge and create a combined chunk
\item Consider related steps or instructions as one inseparable unit of content
\item \textbf{Steps that are part of the same procedure/process must ALWAYS be kept together}
\item \textbf{Even if a set of steps is very long, do NOT split them - they must remain in a single chunk}
\item \textbf{Prioritize keeping steps together over any other chunking considerations}
\item If you encounter steps that seem to be part of the same process but are separated by other content, analyze carefully to determine if they are truly part of the same procedure and should be combined
\end{itemize}

\begin{enumerate}
\setcounter{enumi}{8}
\item Avoid chunks under 3 lines; merge them with adjacent content and heading.

\item Exclude menus, cookie notices, privacy policies, and terms sections.

\item For all heading levels (first, second, and third), ensure complete preservation of details:
    \begin{itemize}
    \item First-level heading: Include full document title, all location details, and audience roles if any.
    \item Second-level heading: Capture complete section names with any qualifying details or descriptions
    \item Third-level heading: Retain all subtopic specifics including numbers, dates, and descriptive text
    \item Never truncate or abbreviate any heading content at any level.
    \end{itemize}

\item Multilingual Support (CRITICAL)
    \begin{itemize}
    \item Multilingual content \textbf{must} be processed with the exact same rules as monolingual content.
    \item Do not skip, paraphrase, or translate non-English content—\textbf{all languages must be preserved and chunked}.
    \end{itemize}

\item MULTI-PAGE CONTEXT HANDLING
\begin{itemize}
\item Ensure contextual continuity between pages during processing
\item When content splits across pages, maintain coherence and proper flow
\item Handle page breaks within paragraphs, lists, or other content blocks seamlessly
\item Track and preserve semantic relationships across page boundaries
\end{itemize}

\item LAYOUT ELEMENTS
\begin{itemize}
\item Remove page headers and footers consistently across all pages
\item Preserve footnotes and endnotes with proper linking to their references
\item Maintain paragraph spacing and indentation
\item Handle multi-column layouts by properly sequencing the content
\item Preserve bulleted and numbered lists with their hierarchical structure
\end{itemize}

\item SPECIAL CONTENT TYPES
\begin{itemize}
\item Process scanned pages with OCR-extracted text while maintaining formatting
\item Preserve the structural integrity of content when images appear within text
\item Extract and describe flowcharts, diagrams, and other visual elements
\item If a Flowchart, describe step by step the flow
\item Extract text from images embedded in the PDF if relevant to surrounding content
\item If the Image is a screenshot, exclude it
\item Include appropriate alt text or descriptions for non-extractable visual elements
\end{itemize}

\item FAQ Separation

When encountering FAQ content, split question-answer pairs into individual chunks rather than grouping them into single large chunks.
\end{enumerate}

\vspace{0.3em}
\textbf{When working with tables:}
\begin{enumerate}
\item Format using proper table syntax (pipes | and hyphens -).
\item Maintain table structure across images if a table spans multiple images.
\item When a table continues from a previous chunk (indicated in LAST CHUNKS), strictly maintain the same column structure, width, and formatting as established in the previous chunk for consistency.
\item \textbf{VERY IMPORTANT}: Create a separate chunk for EACH ROW of the table. Every table row chunk must include the table headers mentioned in the previous chunk or in the image followed by just that single row of data.
\item For each table row chunk, repeat the full table headers to ensure context is maintained independently.
\item If you find a row which is continuing from LAST CHUNKS, continue segmenting without including the content of the previous chunk.
\end{enumerate}

\vspace{0.3em}
\textbf{HOW TO IDENTIFY STEPS AND INSTRUCTIONS:}
\begin{itemize}
\item Look for bulleted lists that describe a process
\item Look for content with clear sequencing words (First, Next, Then, Finally)
\item Look for any content that describes how to perform a task or procedure
\item Look for sections titled "Instructions," "Procedure," "How to," "Guide," etc.
\item Look for multiple paragraphs that clearly belong to the same process
\end{itemize}

\vspace{0.3em}
\textbf{Flag for Content Continuation}

ADD A CONTINUES FLAG TO EACH CHUNK:

For each chunk, you must add a CONTINUES flag:
\begin{itemize}
\item \texttt{[CONTINUES]True[/CONTINUES]}: This chunk is a continuation of the previous chunk OR is part of the same process, instruction set, or procedure.
\item \texttt{[CONTINUES]False[/CONTINUES]}: This chunk starts new content and is not a continuation.
\item \texttt{[CONTINUES]Partial[/CONTINUES]}: This chunk might be related to the previous chunk, but you are not sure.
\end{itemize}

\vspace{0.3em}
\textbf{Flag Rules for Table Rows:}
\begin{itemize}
\item For table row chunks, set the CONTINUES flag specifically as follows:
  \begin{itemize}
  \item \textbf{[CONTINUES]True[/CONTINUES]}: ONLY if the cell content continues from an incomplete cell in the previous chunk/call
  \item \textbf{[CONTINUES]False[/CONTINUES]}: When the row contains complete cell content, NOT continuing from previous chunk
  \item The flag should be based on the CONTENT INSIDE THE CELLS, not on whether the table itself continues
  \end{itemize}
\end{itemize}

\vspace{0.3em}
\textbf{Flag Rules for Steps and Instructions:}
\begin{itemize}
\item For chunks containing numbered steps, instructions, procedures, or lists:
  \begin{itemize}
  \item \textbf{When processing steps/instructions that span multiple pages or images:}
    \begin{itemize}
    \item If steps continue from LAST CHUNKS, use [CONTINUES]True[/CONTINUES]
    \end{itemize}
  \item \textbf{When identifying if steps are complete:}
    \begin{itemize}
    \item Look for clear indications like "Final Step" or concluding language
    \end{itemize}
  \item \textbf{ALL subsequent chunks containing ANY steps from the same procedure MUST use [CONTINUES]True[/CONTINUES]}
  \item The only time a chunk containing steps should use [CONTINUES]False[/CONTINUES] is when it's a completely different procedure with no relation to previous steps
  \end{itemize}
\end{itemize}

\vspace{0.3em}
\textbf{Flag Rules for Other Content:}
\begin{itemize}
\item Use CONTINUES=True for content that directly continues from the previous chunk
\item For general content not falling into the above categories, use your best judgment based on context
\end{itemize}

\vspace{0.3em}
\textbf{Output Requirements:}
\begin{enumerate}
\item Output a list of chunks where each chunk starts with a full 3-level heading and remove all empty or no finding chunks.
\item Use this exact format: 

{\small\texttt{[CONTINUES]True|False|Partial[/CONTINUES][HEAD]main\_heading > section\_heading > \\ chunk\_heading[/HEAD]chunk\_content}}

\item Separate chunks like this: 

{\footnotesize
\begin{verbatim}
[CONTINUES]True|False|Partial[/CONTINUES]
[HEAD]main_heading > section_heading > chunk_heading[/HEAD]
chunk1 

[CONTINUES]True|False|Partial[/CONTINUES]
[HEAD]main_heading > section_heading > chunk_heading[/HEAD]
chunk2
\end{verbatim}
}
\end{enumerate}

\vspace{0.3em}
\textbf{FINAL CHECK BEFORE SUBMITTING:}
\begin{itemize}
\item Have you kept ALL numbered steps together in the same chunk? \textbf{This is critical!}
\item Have you separated FAQ question-answer pairs into individual chunks instead of grouping them together?
\item Have you identified all step sequences correctly and combined them, even if they span multiple pages?
\item Have you identified and skipped all table of contents and indexes?
\item Have you preserved and included all non-English/multilingual content, treating it with the same importance as English?
\item If tables exist, did you follow the special instructions for tables, creating a separate chunk for EACH ROW with headers?
\item Have you applied the correct flag rules for table rows (based on cell content completeness)?
\item Have you kept ALL related procedures together?
\item Have you maintained ALL lists as single units?
\item Have you preserved the integrity of ALL instructional sequences?
\item Have you properly handled content that continues from a previous batch?
\item Have you indicated content that continues to the next batch?
\item Have you added the [CONTINUES] flag to each chunk with appropriate values?
\end{itemize}

\textbf{If you find ANY instances where related steps are split across chunks, recombine them immediately before submitting your final answer.}

Ensure every chunk is clear, fully contextual, and no data is missing.

\vspace{0.5em}
}
\end{mdframed}

\subsection{Prompt for Evaluation}

\begin{mdframed}[
    frametitle={Evaluation Prompt},
    frametitlebackgroundcolor=lightblue!50,
    backgroundcolor=lightblue!10,
    linecolor=lightblue!50,
    linewidth=1pt,
    roundcorner=3pt,
    innertopmargin=10pt,
    innerbottommargin=10pt,
    innerleftmargin=10pt,
    innerrightmargin=10pt
]

{\small
{\itshape Instruction: Read the given Question, Search Results, and Answer. Evaluate whether the knowledge sources contain the necessary information to answer the query and assess the quality of the bot's response.

Provide your output in plain JSON format.

Evaluation Criteria:

is_answer_exist: Determine whether the provided knowledge sources contain information that can be used to answer the user query. Mark True if the knowledge includes content that directly or inferentially answers the question. Mark False if the knowledge does not contain the necessary information, is unrelated, or insufficient to address the query.

response_quality: Assess how the LLM handled the query based on the knowledge provided. Choose one of the following labels:

- "correct" – The model answered the query accurately using relevant information from the knowledge. No unsupported inferences or hallucinations.

- "hallucinate" – The model introduced information not found in or not supported by the knowledge, or it incorrectly assumed relevance from unrelated content.

- "abstain" – The model acknowledged it could not answer due to lack of information, or refrained from answering based on absence of relevant knowledge.

answer_ids: List the specific knowledge source indices or document references used (or that should have been used) to support the LLM's answer. If the answer is not present in the Search Results, return an empty List.
}}
\end{mdframed}

\subsection{Chunk Quality Comparison Examples}

To illustrate the superior quality and structure preservation of our vision-guided chunking approach, we present comparative examples of chunks generated by traditional vanilla chunking versus our multimodal method.

\textbf{Vanilla Chunking Output:}
\begin{mdframed}[
    frametitle={Vanilla RAG Chunk},
    frametitlebackgroundcolor=red!20,
    backgroundcolor=red!5,
    linecolor=red!30,
    linewidth=1pt,
    roundcorner=3pt,
    innertopmargin=8pt,
    innerbottommargin=8pt,
    innerleftmargin=8pt,
    innerrightmargin=8pt
]
{\small
\textbf{Heading:}  Section 36

\textbf{Content:} [If our third-party service providers and business partners do not satisfactorily fulfill their commitments and responsibilities, ourfinancial results could suffer.In the conduct of our business, we rely on relationships with third parties, including cloud data storage and other information technology service providers, suppliers, distributors, contractors, joint venture partners and other external business partners, forcertain services in support of key portions of our operations. These third parties are subject to similar risks as we are relating to cybersecurity, privacy violations, business interruption, and systems and employee failures, and are subject to legal, regulatory and market risks of their own. Our third-party service providers and business partners may not fulfill their respective commitments and responsibilities in a timely manner and in accordance with the agreed-upon terms or applicable laws. In addition, while we have procedures in place for assessing risk along with selecting, managing and monitoring our relationships with third-party service providers and other business partners, we do not have control over their business operations or governance and compliance systems, practices and procedures, which increases our financial, legal, reputational and operational risk. If we are unable to effectively manage our third-party relationships, or for any reason our third-party service providers or business partners fail to satisfactorily fulfill their commitments and responsibilities, our financial results could suffer.If we are unable to renew collective bargaining agreements on satisfactory terms, or if we or our bottling partners experience strikes, work stoppages or labor unrest, our business could suffer. Many of our employees at our key manufacturing locations and bottling plants are covered by collective bargaining agreements.While we generally have been able to renegotiate collective bargaining agreements on satisfactory terms when they expire and regard our relations with employees and their representatives as generally satisfactory, negotiations may nevertheless be challenging, as the Company must have competitive cost structures in each market while meeting the compensation and benefits needs of our employees. If we are unable to renew collective bargaining agreements on satisfactory terms, our labor costs could increase, which could affect our profit margins. In addition, many of our bottling partners’ employees are represented by labor unions. Strikes, work stoppages or other forms of labor unrest at any of our major manufacturing facilities or at our bottling operations or our major bottlers’ plants could impair our ability to supply concentrates and syrups to our bottling partners or our bottlers’ ability to supply finished beverages to customers, which could reduce our net operating revenues and could expose us to customer claims. Furthermore, from time to time we and our bottling partners restructure manufacturing and other operations to improve productivity, which may have negative impacts on employee morale and work performance, result in escalation of grievances and adversely affect the negotiation of collective bargaining agreements. If these labor relations are not effectively managed at the local level, they could escalate in the form of corporate campaigns supported by the labor organizations and could negatively affect our Company’s overall reputation and brand image, which in turn could have a negative impact on our products’ acceptance by consumers. RISKS RELATED TO CONSUMER DEMAND FOR OUR PRODUCTS Obesity and other health-related concerns may reduce demand for some of our products. There is growing concern among consumers, public health professionals and government agencies about the health problems associated with obesity. Increasing public concern about obesity; other health-related public concerns surrounding consumption of sweetened beverages; potential new or increased taxes on sweetened beverages by government entities to reduce consumption or to raise revenue; additional governmental regulations concerning the advertising, marketing, labeling, packaging or sale of our sweetened beverages; and negative publicity resulting from actual or threatened legal actions against us or other companies in our industry relating to the marketing, labeling or sale of sweetened beverages may reduce demand for, or increase the cost of, our sweetened beverages, which could adversely affect our profitability. If we do not address evolving consumer product and shopping preferences, our business could suffer.Consumer product preferences have evolved and continue to evolve as a result of, among other things, health, wellness and nutrition considerations, including concerns regarding caloric intake associated with sweetened beverages and the perceived undesirability of artificial ingredients; concerns regarding the perceived health effects of, or location of origin of, ingredients, raw materials or substances in our products or packaging, including due to the results of third-party studies (whether or not scientifically valid); shifting consumer demographics; changes in consumer tastes and needs coupled with a rapid expansion of beverage options and delivery methods; changes in consumer lifestyles; concerns regarding the environmental, social and sustainability impact of ingredient sources and the product manufacturing process; consumer emphasis on transparency related to ingredients we use in our products and collection and recyclability of, and amount of recycled content contained in, our packaging containers and other materials; concerns about the health and welfare of animals in our dairy supply chain; and competitive product and pricing pressures. In addition, in many of our markets, shopping patterns are being affected by the digital evolution, with consumers rapidly embracing shopping by way of mobile device applications, e-commerce retailers and e-commerce websites or platforms. If we fail to address changes in consumer product and shopping preferences, do not successfully anticipate and prepare for future changes in such preferences, or are ineffective or slow in developing and implementing appropriate digital transformation initiatives, our share of sales, revenue growth and overall financial results could be negatively affected.04/01/2025, 16:40 ko-20221231https://www.sec.gov/Archives/edgar/data/21344/000002134423000011/ko-20221231.htm 22/170]
\\
\\
\\
}
\end{mdframed}

\textbf{Vision-Guided Chunking Output:}
\begin{mdframed}[
    frametitle={Vision-Guided RAG Chunk},
    frametitlebackgroundcolor=green!20,
    backgroundcolor=green!5,
    linecolor=green!30,
    linewidth=1pt,
    roundcorner=3pt,
    innertopmargin=8pt,
    innerbottommargin=8pt,
    innerleftmargin=8pt,
    innerrightmargin=8pt
]
{\small
\textbf{Heading:} ko-20221231 > Part I > ITEM 1A. RISKS RELATED TO CONSUMER DEMAND FOR OUR PRODUCTS>

\textbf{Content:} [If we do not address evolving consumer product and shopping preferences, our business could suffer. Consumer product preferences have evolved and continue to evolve as a result of, among other things, health, wellness and nutrition considerations, including concerns regarding caloric intake associated with sweetened beverages and the perceived undesirability of artificial ingredients; concerns regarding the perceived health effects of, or location of origin of, ingredients, raw materials or substances in our products or packaging, including due to the results of third-party studies (whether or not scientifically valid); shifting consumer demographics; changes in consumer tastes and needs coupled with a rapid expansion of beverage options and delivery methods; changes in consumer lifestyles; concerns regarding the environmental, social and sustainability impact of ingredient sources and the product manufacturing process; consumer emphasis on transparency related to ingredients we use in our products and collection and recyclability of, and amount of recycled content contained in, our packaging containers and other materials; concerns about the health and welfare of animals in our dairy supply chain; and competitive product and pricing pressures. In addition, in many of our markets, shopping patterns are being affected by the digital evolution, with consumers rapidly embracing shopping by way of mobile device applications, e-commerce retailers and e-commerce websites or platforms. If we fail to address changes in consumer product and shopping preferences, do not successfully anticipate and prepare for future changes in such preferences, or are ineffective or slow in developing and implementing appropriate digital transformation initiatives, our share of sales, revenue growth and overall financial results could be negatively affected.]
}
\end{mdframed}

\end{document}